\documentclass{article}

\usepackage{float}
\usepackage{PRIMEarxiv}
\usepackage{amsmath} 
\usepackage{authblk}
\usepackage[utf8]{inputenc} % allow utf-8 input
\usepackage[T1]{fontenc}    % use 8-bit T1 fonts
\usepackage{hyperref}       % hyperlinks
\usepackage{url}            % simple URL typesetting
\usepackage{booktabs}       % professional-quality tables
\usepackage{amsfonts}       % blackboard math symbols
\usepackage{nicefrac}       % compact symbols for 1/2, etc.
\usepackage{microtype}      % microtypography
\usepackage{lipsum}
\usepackage{fancyhdr}       % header
\usepackage{graphicx}       % graphics
\graphicspath{{media/}}     % organize your images and other figures under media/ folder

\usepackage{bbm}

\title{Automated Remote Sensing Forest Inventory Using Satellite Imagery}
\author[a,*]{Abduragim Shtanchaev}
\author[a,b]{Artur Bille} 
\author[a]{Olga Sutyrina}
\author[a]{Sara Elelimy}
\affil[a]{Skolkovo Institute of Science and Technology, Bolshoy Boulevard 30, bld. 1, Moscow, Russia}

\affil[b]{Ulm University, Helmholtzstraße 18, Ulm, Germany}
\affil[*]{Corresponding Author}
\begin{document}
\insert\footins{  \noindent Abduragim Shtanchaev \hspace*{3.5cm} Artur Bille \\ 
\texttt{abduragim.shtanchaev@skoltech.ru} \hspace*{1.15cm} \texttt{artur.bille@uni-ulm.de}  \vspace*{0.2cm} \\  
Olga Sutyrina \hspace*{4.7cm} Sara Elelimy\\ 
\texttt{olga.sutyrina@skoltech.ru} \hspace*{2.3cm} \texttt{sara.elelimy@skoltech.ru}}
\maketitle

\begin{abstract}
% \lipsum[1]
For many countries like Russia, Canada, or the USA, a robust and detailed tree species inventory is essential to manage their forests sustainably. Since one can not apply unmanned aerial vehicle (UAV) imagery-based approaches to large scale forest inventory applications, the utilization of machine learning algorithms on satellite imagery is a rising topic of research. Although satellite imagery quality is relatively low, additional spectral channels provide a sufficient amount of information for tree crown classification tasks. Assuming that tree crowns are detected already, we use embeddings of tree crowns generated by Autoencoders as a data set to train classical Machine Learning algorithms. We compare our Autoencoder (AE) based approach to traditional convolutional neural networks (CNN) end-to-end classifiers.
\end{abstract}

% keywords can be removed
\keywords{Remote Sensing \and Forest Inventory \and 
Semi-Supervised Learning \and Classification}

\section{Introduction}

In the face of environmental changes, including forest deterioration, the demand for a cheap, robust, and rapid automatic forest inventory has significantly increased. It is notably relevant in countries primarily covered with forests such as Russia, Canada, and Brazil. Currently, the most widely employed forest inventory techniques are manual and UAV based approaches. They are time inefficient and expensive. An alternative for those methods emerges from remote sensing. Using widely and readily available satellite imagery can effectively reduce cost and time for forest inventory analysis. We propose a novel, semi-supervised learning method for forest inventory using satellite imagery that addresses satellite imagery's primary challenge - the low quality of images. Any image-based forest inventory generally consists of two elements - detection and classification. This work only investigates the tree spices classification part. Hence the detection is a prerequisite and assumed.

In general, there are two types of land cover classification. The first type is pixel-based, and the second one is an object-based classification. The main difference between these approaches is that the latter treats many pixels corresponding to an object as a whole, whereas the former analyses pixels individually (semantic segmentation). According to Heinzel et al. \cite{heinzel_11}, \cite{heinzel_12}, the object-based shows better results in practice. Following this result, we consider only object-based classification in our work.

In our work, we consider 4 types of tree species for classification: \textit{Birch (Betula fruticosa), Spruce (Picea omorika), Pine (Pinus sibirica), Fir (Abies sibirica)} and at the same time, fight two main challenges of forest inventory classification. The first challenge - the quality of satellite imagery is relatively low for individual tree crow classification. The second challenge is that the data set is extremely small, only 1012 images with tree species labels.  

% \comment{
%     The object-based approach and the 8 channels of each image in our data set, lead to a high number of features for every tree. Therefore, the question arises whether all these features are necessary for accurate tree classification. Another worthy question is if one can reduce the data size without decreasing the quality of classification methods too much. Addressing these questions, we used two types of autoencoders to zip the high dimensional data. Then, we trained well-known machine learning classification methods with the new compressed data set. Further, we applied two convolutional networks, each of a different kind, for an end-to-end classification.
%     The main challenge of the classification task is the big area of $46\times 46~\text{cm}^2$ covered by one pixel, in contrast to $5.5\times 5.5~\text{cm}^2$ per pixel in hyperspectral data.
% }

% This paper is structured as follows. In Section \ref{data_set}, we give a detailed description of the data set. After that, we provide a short overview of related literature dealing with similar problems and using similar strategies. In Section \ref{algo_and_models}, we discuss the details of the explicit algorithms and models of the considered AEs and CNNs. 
% Next, we present experiments and their main results in Section \ref{exp_and_res}. The last Section \ref{conclusion} is devoted to a conclusion and an outlook for further investigations.  

\section{Data set}
\label{data_set}
\begin{figure}[!htbp]
\centering
\includegraphics[width=0.6\columnwidth]{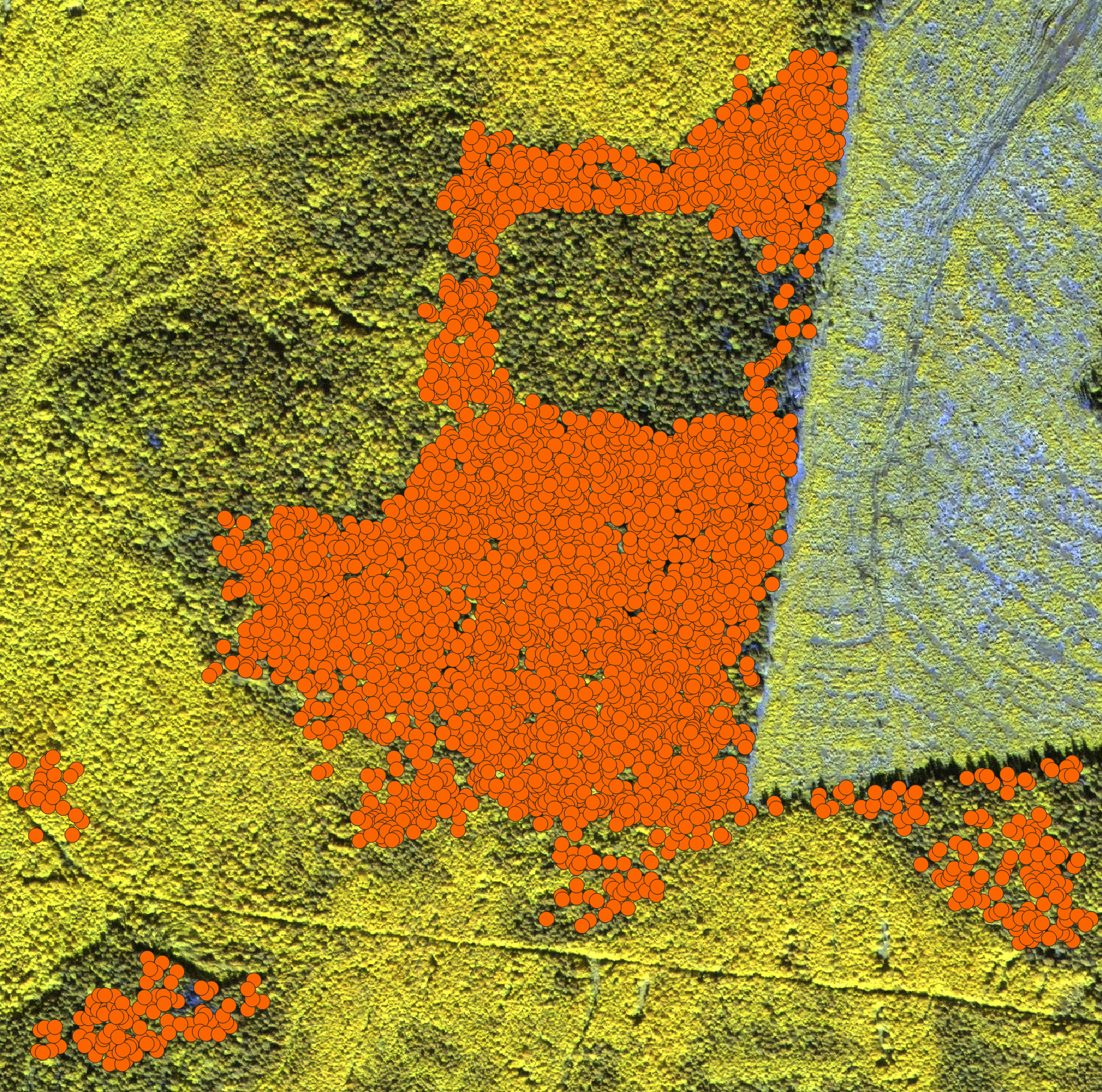}
\caption{\label{coordinats}Ten thousand tree crown coordinates of boreal forests} 
\end{figure}

As this work is a logical continuation of \cite{inproceedings}, we analyze data of the same forest. The authors of \cite{inproceedings} acquired labeled data in the Arkhangelsk region (northern Russia), Krasnoborsky district forestry division, which is covered by plain cultivated boreal forests of high density(900 - 1000 trees/hectare). UAVs gathered the data using hyperspectral and Lidar sensors. In addition to the data set used in \cite{inproceedings}, we created another data set consisting of ten thousand locations of tree crowns in neighboring regions for which no labels exist. This enhanced data set is the base of our semi-supervised learning approach. For the sake of convenience, we are going to refer to the new data set as "unlabeled" data set and to the original data set as "labeled". Figure \ref{coordinats} gives an impression of the additional data.

\begin{figure}
\centering
\includegraphics[width=120pt,angle=90]{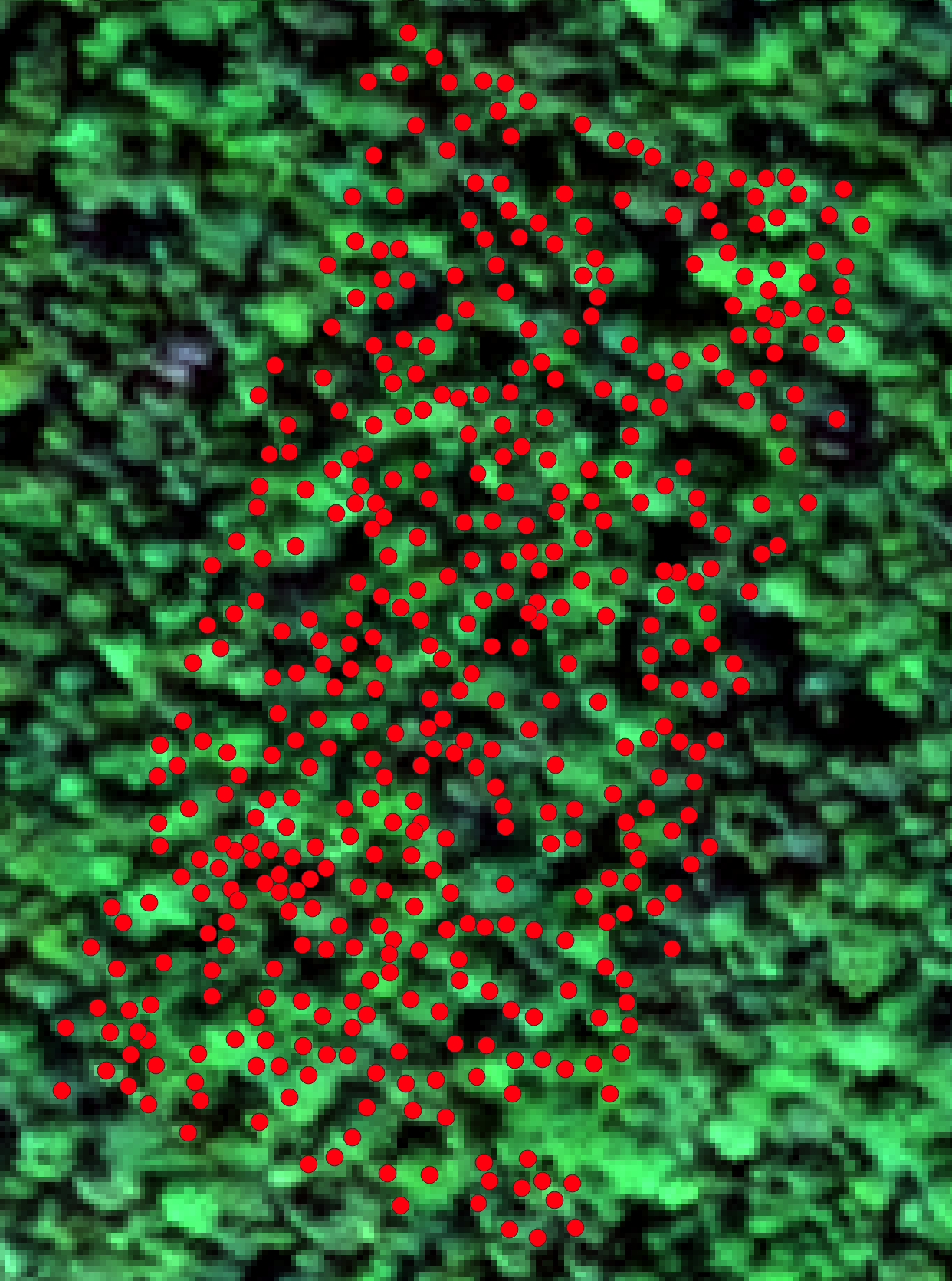}
\caption{\label{labeleddataset}One of four polygons with coordinates and labels of tree crowns. Data from these polygons is used for ML model training and testing.} 
\end{figure}

Authors of \cite{inproceedings} selected four test polygons of size $150 \times 200$m for experimental data collection. The total number of labeled trees is 1012. There are four types of trees: Spruce (Label 0), Birch (Label 1), Fir (Label 2), and Pine (Label 3). Figure \ref{labeleddataset} presents an extraction of the test polygon. This data set is highly imbalanced. The ratios of the four classes are 51.19\% (518 trees), 31.13\% (315 trees), 4.15\% (42 trees), and 13.54\% (137 trees). The data set used in \cite{inproceedings} consists of 32 hyperspectral bands and was acquired using a drone, as mentioned earlier.

Since this work's objective is to use satellite imagery, we had to georeference labels from \cite{inproceedings} data set to WorldView-2 satellite imagery. We preformed georeferencing using Qgis software. This data set consists of trees' locations on WorldView-2 imagery. A sample consists of eight multispectral channels and one high-resolution panchromatic channel of 450 - 800 nm. Four out of the eight channels are standard colors(red, green, blue, near-infrared 1), and four other bands are coastal 400 - 450 nm, yellow 585 - 625 nm, red edge  705 - 745 nm, and near-infrared 2860 - 1040 nm. Multispectral channels have a resolution of 1.84 m$^2$/pixel, whereas panchromatic channels cover 0.46 m$^2$/pixel. 

\section{Algorithms and Models}
\label{algo_and_models}

In this work, we employ several methodologies to tackle the problem of individual tree crown
classification. Classical computer vision techniques use features generated by ``static'' 
algorithms, meaning that such algorithms do not learn to extract the most relevant features for 
a particular test. These features can be of any kind, starting with the mean of image pixels and ending by edges of objects on an image. Neural Networks, on the other hand, offer data driven 
feature generation, which allows selection of the most convenient features, which means that networks can learn to extract the most relevant features.

\begin{figure}[!htbp]
\centering
% \hspace*{-0.6cm}
\includegraphics[width=7in]{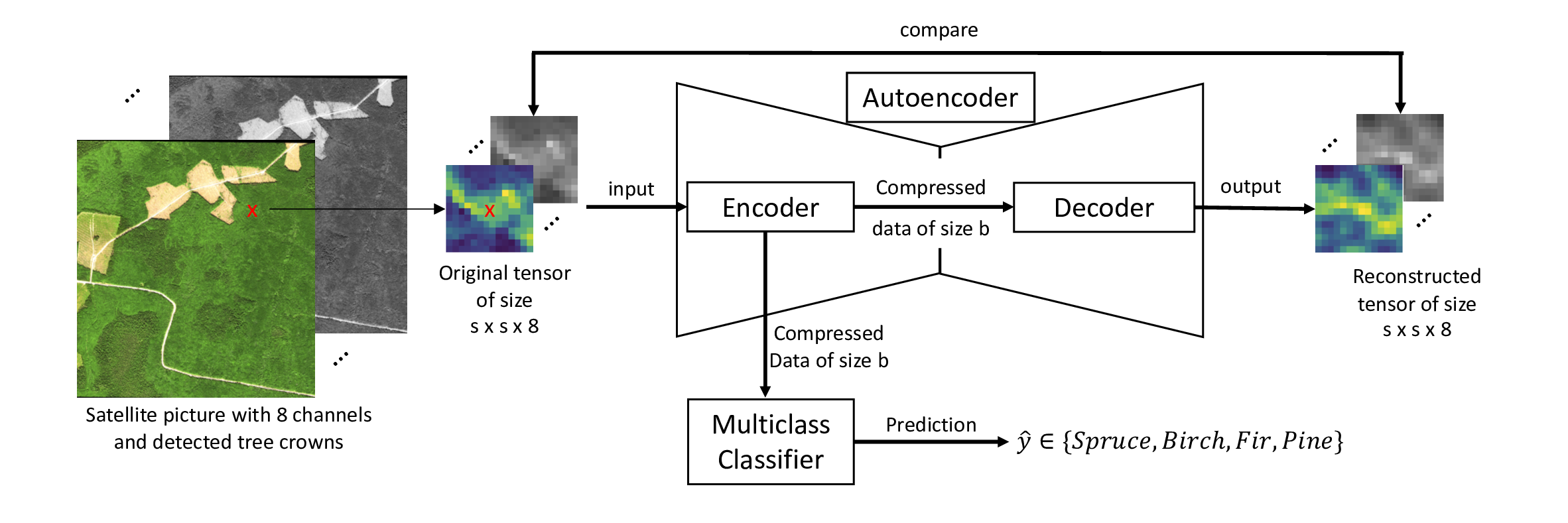}
\caption{\label{icml-historical}Illustration of the general approach. Here, $s^2$ is the number of pixels in every channel in the surrounding of the considered tree crown, marked with a red cross.} 
\end{figure}

In general, a CNN requires high volumes of labeled data for supervised learning. Since our data set is too small, we decided to use AEs to be trained using unlabeled data (that is in abundance in our case). After training AE on the unlabeled data set, we use the encoder to compress the labeled data set into embeddings. These embeddings are the input for the classification methods. The general scheme is illustrated in Figure \ref{icml-historical}. Alongside the AE, we also use CNN for end-to-end classification using the labeled data only(for comparison with AE approach). 

As described in the introduction, every tree's data size depends on the number of pixels considered around the tree crown. To decide whether more information of the surroundings leads to better classification results, we analyzed images of size $s \times s$ with $s\in\lbrace 10+4*k~|~k=0,\ldots,5\rbrace$. Due to the satellite pictures' eight channels, one has a tensor of the size of $s\times s \times 8$ for every tree. Hence, the size of the data set overgrows with the image size.
 
Another critical parameter in our models is the size we want to shrink the data (embedding space). This size is the bottleneck size of our AE, and we consider the following values $\lbrace 40,60,80,100,120 \rbrace$. 
In our work, we compare two different kinds of AEs. The first one uses convolutional layers combined with linear layers, whereas the second one consists of linear layers only.

\subsection{Autoencoders}
\subsubsection{Convolutional autoencoder}
One of the employed approaches is an AE utilizing convolutional layers in its architecture. This choice is motivated by the fact that these layers are well known for extracting the most suitable features out of the images by introducing inductive bias and permeation equivariant. 
The encoder's general structure consists of two convolutional layers followed by two linear layers, and after each layer, we inserted a bach normalizing layer as well as ReLU non-linearity. The number of filters in the first convolutional layer is 60 and in the second 200.

% \comment{
%     \begin{figure}[htbp]
%     \centering
%     \includesvg[width=\columnwidth]{encoder}
%     \caption{Encoder with two convolutional followed by two linear layers. The decoder consists of three linear layers. The tensors are flattened in the encoder and unflattened in the decoder.}
%     \label{cov-encoder}
%     \end{figure}
% }

Here, the AE decoder consists of two linear layers followed by batch normalization and ReLU non-linearity and a sigmoid function at the end.

\subsubsection{Sparse autoencoder}
In contrast to the previous subsection model, we construct an AE using a neural network with three linear hidden layers, which of the second one is the bottleneck with $b$ nodes. Both the first and last layers have $ 0.5\cdot(\text{inputsize}+b)$ nodes, where the input and output of our neural network are of the same size. Since we use vectorized tensors as input, their size is equal to $8s^2$. After training this neural network with unlabeled data, the first two layers are used as an encoder to transform the original data into the latent space of a much lower dimension. Then, we use compressed data as inputs for our classifiers. 
We add the coordinates of trees as two additional elements to the input vector in a slight modification. The basic idea is to investigate whether trees of the same kind are distributed uniformly or concentrated on forests' specific areas. Note that pixels and tree crowns' coordinates are normalized and lie all in $[0,1]$.

\subsection{Convolutional Neural Network}
A CNN is a deep learning algorithm that achieves a high-performance level on a wide range of image classification and computer vision issues. These algorithms have improved cutting-edge technology significantly in image recognition and visual object recognition. Such a network has two essential phases: feature selection and classification. In this section, we are going to discuss two CNN models that we have applied in end-to-end fashion to labeled dataset. One is 2D-CNN with custom architecture, and the other one is a VGG neural network introduced in \cite{Simonyan15}.

\subsubsection{2D-CNN}
\begin{figure}[!htbp]
\centering
\includegraphics[width=0.5\columnwidth]{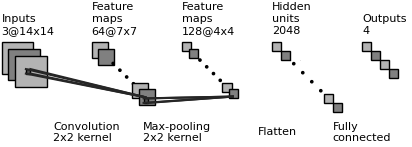}
\caption{CNN architecture with input size $3\times 14\times 14$ \label{fig46}}
\end{figure}

The custom architecture used in 2D-CNN is illustrated in Figure \ref{fig46}. It consists of two convolutional layers. The first and the second layer consists of 68 and 128 filters, respectively. Each convolutional layer has a kernel of size $2\times 2$ followed by a max-pooling layer with the same kernel size. The resulting feature map is then flattened and passed into two fully connected layers that generate a vector four logits for each data instance. All activations are ReLU functions. We also used dropout layers and weighted cross-entropy loss to reduce overfitting and account class imbalanced.

\subsubsection{VGG network}
\begin{figure}[!tbp]
  \centering
  \begin{minipage}[b]{0.4\textwidth}
    \includegraphics[width=\textwidth]{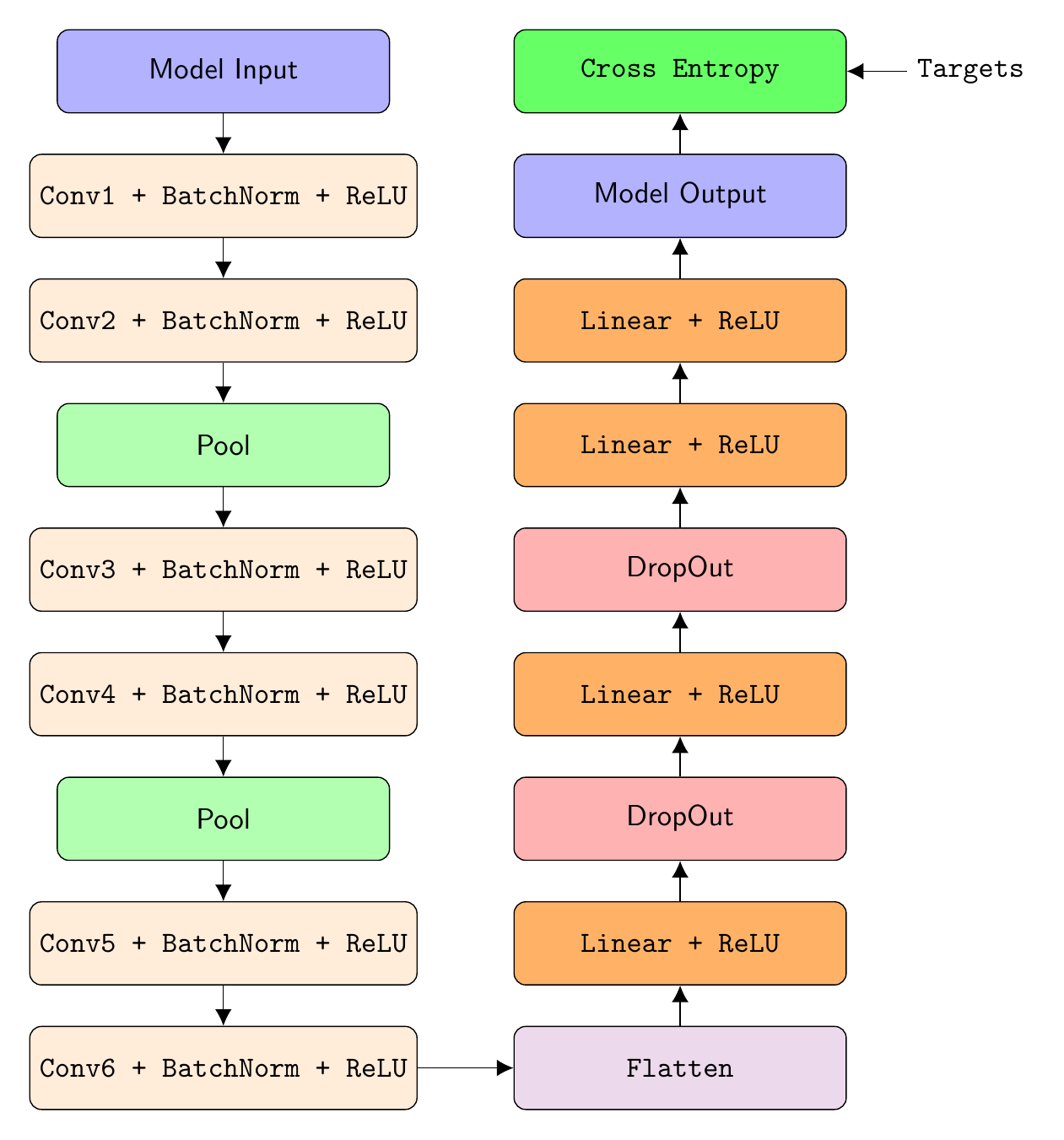}
    \caption{Architecture of the used VGG.}
    \label{VGG}
  \end{minipage}
  \hfill
  \begin{minipage}[b]{0.4\textwidth}
    \includegraphics[width=\textwidth]{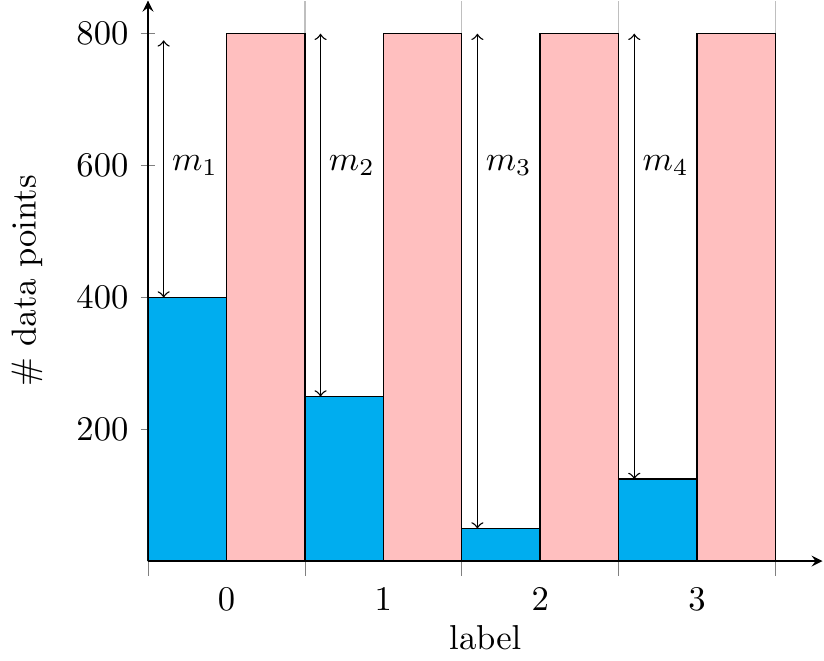}
    \caption{Illustration of the images' duplication process in the train set. Blue and pink bars represent the number of data points with the corresponding label in the unbalanced and upsampled data set, respectively.}
    \label{Dublication}
  \end{minipage}
\end{figure}

As an experiment, we consider the VGG network to see if a very deep, convolutional network could significantly increase data classification accuracy. The architecture of the network is shown in Figure \ref{VGG}. Interested readers can find more detailed information about the used VGG network parameters in the \ref{Vgg_net}. Since the labeled dataset is unbalanced and has a small number of images, we decided to artificially extend it by duplicating the images in the training dataset according to the proportions of the total number of labels to each class's number of labels of the dataset. The duplication process is demonstrated in Figure \ref{Dublication}.

% \begin{figure}[!htbp]
%     \begin{minipage}
%     \includegraphics[width=4cm]{Figures/vgg_architecture.pdf}
%     \includegraphics[width=4cm]{Figures/vgg_architecture.pdf}
%     \caption{Architecture of the used VGG.}
%     \label{VGG}
%     \end{minipage}
% \end{figure}

    % \begin{subfigure}
    % \centerline
    % \includegraphics[width=0.3\columnwidth]{Figures/upsampling.pdf}
    % \caption{Illustration of the images' duplication process in the train set. Blue and pink bars represent the number of data points with the corresponding label in the unbalanced and upsampled data set, respectively.}
    % \label{Dublication}
    % \end{subfigure}

The augmentation is applied to each image when forming batches by a random rotation, shear and translation within the specified range. Formation of "dark" corners during rotating images was prevented by increasing the size of an input image by cropping two times. Note that cropping was also applied for images in the test dataset.

% \begin{figure}[!htbp]
% \begin{center}
% \centerline{\includegraphics[width=0.3\columnwidth]{Figures/upsampling.pdf}}
% \caption{Illustration of the images' duplication process in the train set. Blue and pink bars represent the number of data points with the corresponding label in the unbalanced and upsampled data set, respectively.}
% \label{Dublication}
% \end{center}
% \end{figure}
 
\section{Experiments and Results}
\label{exp_and_res}
All results, code, data sets, and other technical details can found on \href{https://github.com/arturbille/ML-project-Group9-WeakLearners}{GitHub}.

Unless otherwise stated, the following methods and ranges for parameters hold for all succeeding experiments. Additional standard parameters are mentioned in the corresponding subsections. 
We consider image sizes $s$ in $\lbrace 10,14,18,22,26,30 \rbrace$ and bs $b$ in $\lbrace 40,60,80,100,120 \rbrace$. Consistent to \cite{inproceedings}, we utilize (overall) accuracy and F1 score to evaluate the performance of our classifiers. We split the data set in 80\% training data and 20\% test data. The labeled data set consists of images, which have been cut from the satellite image and normalized relative to the maximal pixel value.

\subsection{Autoencoders}
\label{exp_and_res_AE}
This section discusses the results obtained using AEs as feature extractors and a set of classifiers to classify trees. In the following two experiments, we have trained 30 AEs, each using MSE as loss-function. For every image size, we have used all bottleneck sizes mentioned in Section \ref{exp_and_res} and the following classifiers: 

\begin{itemize}
    \item Adaboost, 
    \item XGboost, 
    \item CatBoost (CatB.), 
    \item RandomForest, 
    \item Support Vector Machine (SVC),
    \item Logistic Regression.
\end{itemize}

% The first AE was trained for 300 epochs, the second one for 100 epochs. 

\subsubsection{Convolutional Autoencoder}

In the first experiment, we consider features generated by AEs based on convolutional layers. Each model was trained six times, each with a different image size using the unlabeled data set of 10,000 images with corresponding image sizes mentioned above. To match sizes of labeled images with those images on which AE was trained, we cropped little “patches” of the same size from satellite images according to a tree crown's location. Once labeled images of the same size were obtained, the trained AE encoder was used to convert all labeled images to the embedding space. Then embeddings were used for classification as we know labels for each embedding. The table \ref{F1table} below shows the best results(accuracy and f1 score) of classification for each bottleneck and image size. We only included the best classifier results (SVC in this case), and the overall best result is in bold text.

\begin{table}[!htbp]
    \centering
    \caption{The accuracy and F1-scores of the best classifiers for different bottleneck values and fixed image size $10\times 10$. The first column lists all considered bottleneck sizes.
    \label{F1table}}
    % \label{tab:my_label}
    \begin{tabular}{|c|c|c|}\hline
        & \begin{tabular}{c}$10\times 10$\\  {Accuracy} \end{tabular} & \begin{tabular}{c}$10\times 10$\\   F1-score \end{tabular} \\
        \hline
         40 & \begin{tabular}{c}   SVC:\\   0.964/0.753 \end{tabular} &\begin{tabular}{c}   SVC:\\   0.964/0.636 \end{tabular} \\
         \hline
         60 & \begin{tabular}{c}   \textbf{SVC:}\\   \textbf{0.971/ 0.768} \end{tabular} &\begin{tabular}{c}   \textbf{SVC:}\\    \textbf{0.97/0.645} \end{tabular} \\
         \hline

         80 & \begin{tabular}{c}   SVC: \\   0.974/0.743 \end{tabular} & \begin{tabular}{c}   SVC: \\   0.974/0.631 \end{tabular} \\
         \hline
         
         100 & \begin{tabular}{c}   CatB..: \\   0.999/0.763 \end{tabular} &\begin{tabular}{c}   SVC: \\   0.969/0.639 \end{tabular} \\
         \hline
         120 & \begin{tabular}{c}   SVC: \\   0.977/0.768 \end{tabular} &\begin{tabular}{c}   SVC: \\   0.977/0.645 \end{tabular}\\
         \hline
         
    \end{tabular}
\end{table}

From Table \ref{F1table}, one can observe that almost unanimously, SVC with RBF kernel gives better results than all other classifiers we have considered in this experiment. Secondly, there is a large overfitting happening in the accuracy metric and the f1 score metric. In Section \ref{conclusion}, the overfitting issue is discussed in more detail. One should note that the embeddings were up-sampled using a random oversampling method for the training set, and the test set was left as it is, since tree classes were initially too imbalanced.
From this experiment, one can conclude that using wide image sizes increase neither accuracy nor f1 score. Moreover, sometimes large image size negatively affects the predictive power of a model. Thus only a single tree without its neighborhood should be used for classification. Also, a bottleneck of 60 neurons gave the best accuracy and f1 score. It shows that a bottleneck with more neurons retains useless information that confuses the model.

\subsubsection{Neural Network Autoencoders}
\label{SAE}
We ran this experiment twice, with and without normalized coordinates of tree crowns as additional features.
First, let us shortly discuss the MSE between the original vectorized image and the reconstructed vector generated by our neural network. 
Since for increasing image size $s$, the MSE is increasing automatically, we fix $s$ and observe that increasing $b$ leads mostly to smaller MSE-results( see Figure \ref{AE_MSE_s5} in Appendix). On the test set, one can see the MSE decreasing by 30 - 40\% if one increases $b$ from 40 to 120 in both cases, with and without coordinates. The least overfitted model, i.e., the one with the smallest difference between MSE on training and test set, is given for size image $10\times 10$. For all bottleneck sizes, the MSE of the test set is about 7\% to 10\% higher than on the training set. All other combinations of image size and bottleneck size lead to an MSE gap of at least 15\% and up to 140\%.

\begin{table}[!htbp]
\centering\caption{Highest achieved accuracy and f1 scores of SAE with subsequent classifier on test set with/without using coordinates. The listed bottleneck size $b$ and classifier are those which yield the highest scores.  \label{sparse_ae_best_resuls1}}
\begin{tabular}{|l|c|c|c|}\hline
  Image size &  $10\times10$ &  $14\times 14$ &  $18\times 18$\\ 
\hline 
  Accuracy &   \textbf{0.739/0.739} &   0.734/0.734 &   0.7/0.709\\
$b$ &   \textbf{60/40} &   40/80 &   80/100 \\
  classifier &   \textbf{CatB./CatB.} &   SVC/CatB. &   SVC/XGB \\
\hline
  F1-score &   \textbf{0.62/0.62} &   0.61/0.608 &   0.538/0.543 \\
$b$ &  \textbf{100/40} &   100/80 &   80/120 \\
  classifier &  \textbf{CatB./CatB.} &   CatB./CatB. &   SVC/XGB\\
\hline \hline
  Image size &   $22\times 22$ &   $26\times 26$ &   $30\times 30$\\
\hline
  Accuracy &   \textbf{0.69/0.695} &   0.67/0.665 &   0.67/0.665 \\
$b$ &   \textbf{60/40} &   40/80 &   80/100 \\
  classifier &   \textbf{CatB./CatB.} &   SVC/CatB. &   SVC/XGB\\
\hline
  F1-score &   0.555/0.528 &   \textbf{0.564/0.552} &   0.513/0.513 \\
$b$ &   60/120 &   \textbf{120/100} &   80/120\\
  classifier &   CatB./CatB. &   \textbf{SVC/CatB.} &   CatB./XGB\\ \hline
\end{tabular}
\end{table}

Next, let us discuss the best results of the trained classifiers w.r.t. accuracy and f1 score. In general, comparing any of these scores on train and test data shows that these models are overfitted. Nevertheless, one can conclude tendencies which parameters and which methods work the best on the given data set. As shown in Table \ref{sparse_ae_best_resuls1}, the overall best accuracy occurred using the smallest image size and a relatively small bottleneck size. Further, the coordinates' usecoordinates' use in this experiment did not influence the classifier's accuracy and f1 score. The following observations may explain this. First, the amount of labeled data, even after resampling the minority classes, is not sufficient to train classifiers properly. Secondly, the different kinds of trees seem to be distributed arbitrarily. There are no noteworthy areas on the satellite picture which contain one species only, such that one could neglect the position of the considered tree. Last, it might be a methodology problem. Normalizing the coordinates and treating them like pixels in the AE and the classification methods might dilute this information at all. The latter argument can be analyzed in a future experiment using models with several kinds of inputs. 

In Table \ref{sparse_ae_best_resuls1}, one observes that Adaboost and logistic regression both did not perform best in any scenario w.r.t. accuracy and F1-score, which leads to the idea that the other classifiers suit this problem better. In particular, either CatB. or SVC has the highest accuracy in most cases. Note that each time XGboost performed best, we considered coordinates as additional features. 

Another observation is that the smaller the image size, the higher is the accuracy and f1 score, see Table \ref{sparse_ae_best_resuls1}. Contrary to our intuition, a smaller bottleneck size yields better accuracy results in many cases. Exploring the data in more detail implies that increasing the number of features in the embedding space does increase the computational costs, but neither the accuracy nor the F1-score.

\subsection{2D-CNN}
\label{2d_CNN}

This section presents the results of experiments produced by shallow 2D CNN with custom architecture. As we have mentioned, we would like to compare shallow CNNs with deep CNNs.

\subsubsection{Preprocessing Techniques and Methods }
\label{Preprocessing Techniques and Methods}
Data augmentation simple techniques that revealed remarkable results, as shown in many recent research work \cite{wang2017effectiveness}.
We have applied all the following transformations with the probability 0.5 to our images: horizontal flip, random contrast, random gamma, random brightness, random rotation of 90$^{\circ}$, transpose, shift, scale, blur, optical distortion, and grid distortion. 

\subsubsection{ Training Parameters}
\label{Model Description and Training Parameters}   
For this experiment, the authors used two loss functions: first, a weighted cross-entropy loss function with weights based on the below formula to overcome ``label  0”'s minor accuracy. Second, the same loss function but equal weights.

\begin{equation}\label{eq44}
    w_{t} = \frac{\sum_{i=0}^{n} \mathbbm{1}_{i=t} }{\sum_{i=0}^{n} \mathbbm{1}}
\end{equation}

 Adam optimizer is selected to update the model's parameters with a learning rate set to $10^{-3}$. 
 
\begin{figure}[!htbp]
\centering
\includegraphics[width=0.5\linewidth]{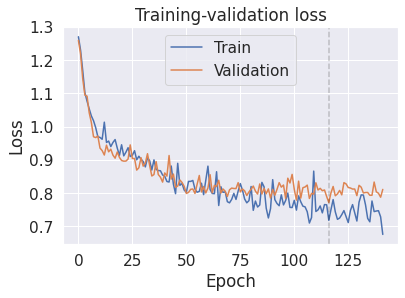}
\caption{Training - Validation loss vs Epochs.   \label{fig42}}
\end{figure}

  Furthermore, an early-stopping-technique is used for the training model, as illustrated in Figure~\ref{fig42}. The dashed line indicates the early-stopping epoch number.

\subsubsection{Model Evaluation and Results}
\label{Model Evaluation and Results}

\begin{figure}[!htbp]
\centering
\includegraphics[width=0.8\linewidth]{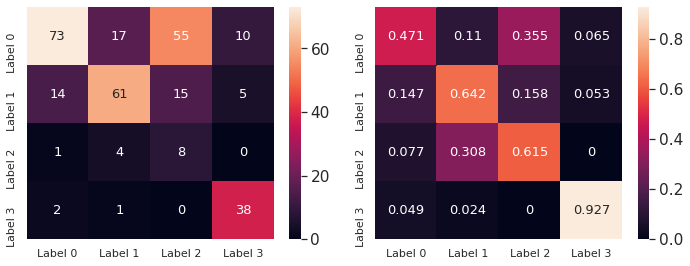}
\caption{Confusion Matrix of the CNN  \label{fig43}}
\end{figure}

Results in the form of a confusion matrix for the model using $14\times 14$ image size are presented in Figure~\ref{fig43}. The considered models achieved the highest accuracy for Pine tree crowns class along with other classes. The lowest classification accuracy was measured for the fir tree class; whereas identifying spruces and birches, average classification accuracy was achieved. The Receiver Operating Curve (ROC) of the CNN model on all data layers with area under curve (AUC) values for each class can be found in Appendix Figure~\ref{fig44}.

We measured the error of the presented solution using a weighted cross-entropy loss function approach. By that, the usual and upgraded CNN achieved a satisfying accuracy for the Fir tree class. Using image size $14 \times 14$ achieved the highest overall accuracy and F1-score of almost 0.6 and 0.58, respectively, compared to other used image sizes listed in Table~\ref{acc2}.

\begin{table}[!htbp]
\centering\caption{Accuracy and F1-score of 2D-CNN and upgraded 2D-CNN for different image sizes.\label{acc2}}
\begin{tabular}{|@{}c@{}|c|c|}
\hline
      &   CNN &   Upgraded CNN  \\
     \hline
     \begin{tabular}{c}   Image size\\
      $10\times 10$\\   $14\times 14$ \\   $18\times 18$ \\   $22\times 22$ \\   $26\times 26$ \\   $30\times 30$
    \end{tabular} & 
    \begin{tabular}{@{}c|c@{}}   Accuracy &   F1 \\    0.563 &   0.55 \\   \textbf{0.603} &   \textbf{0.576} \\   0.579 &   0.563 \\   0.586 &   0.548 \\   0.607 &   0.56 \\   0.559 &   0.53\\ 
    \end{tabular} & 
    \begin{tabular}{@{}c|c@{}}   Accuracy &   F1\\    0.758 &   0.664 \\  \textbf{0.773} &  \textbf{0.656} \\   0.758 &   0.674 \\   0.758 &   0.681 \\   0.758 &   0.634 \\   0.763 &   0.695 \\ \end{tabular}\\\hline
\end{tabular}
\end{table}

Adding convolutions, balancing the training dataset, and the "flip" augmentation procedure can help to achieve an increase in accuracy. The detailed architecture is represented in the \ref{custom_net_}.

\subsection{VGG network}
\label{vgg_network}
The VGG model has been tested on images of following sizes $10\times 10$, $14\times 14$, $18\times 18$, $22\times 22$, $26\times 26$, $30\times 30$. The train dataset contains 809 images, the test dataset contains 203 images.

\begin{figure}[!htbp]
\begin{center}
\centerline{\includegraphics[width=0.86\columnwidth]{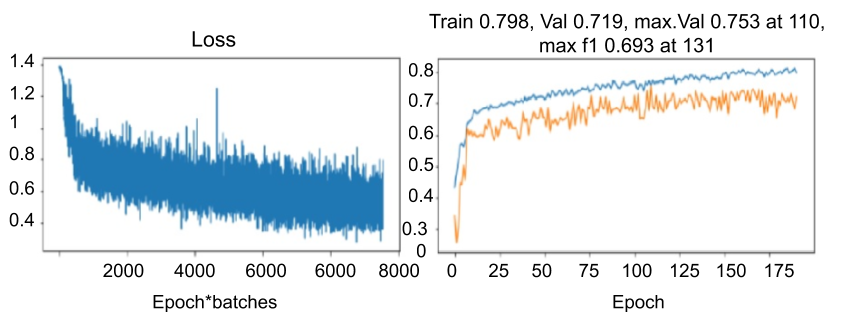}}
\caption{Loss function and dynamic of accuracy metric for VGG}
\label{Loss_acc_VGG}
\end{center}
\end{figure}

One estimates the accuracy of classification algorithms by two metrics: f1 score and accuracy. The final results of both metrics are presented in Table \ref{table3}. 
The dynamics of the loss function for the train set and accuracy metric change is shown in Figure \ref{Loss_acc_VGG} for $10\times 10$ images. The evaluation of the loss function was performed on every batch of each epoch (200 epochs).

\begin{table}[H]
\centering\caption{Accuracy and F1-score using VGG network and different image sizes. \label{table3}}
\begin{tabular}{|c|c|c|}
\hline
Image size &Accuracy & F1\\ 
\hline
$10\times10$ &   \textbf{0.75} &   \textbf{0.69}\\
  $14\times 14$ &   0.7 &   0.64\\
  $18\times 18$ &   0.67 &   0.62 \\
  $22\times 22$ &   0.67 &   0.62\\
  $26\times 26$ &   0.64 &   0.6\\
  $30\times 30$ &   0.6 &   0.57\\ \hline
\end{tabular}
\end{table}

Our results show that the model yields the best results with the image size of $10\times 10$. One explanation of this outcome could be the fact that several tree crowns can be on one image. This experiment demonstrates that deep convolutional networks perform better than external custom networks. Also, the results of the experiment demonstrate that, for the given data set, supervised end-to-end learning is only slightly worse than the semi-supervised learning method employed in the experiments above. Further research with pre-trained deep networks should be conduct as this experiment leaves great promising. 

\section{Conclusion}
\label{conclusion}
All in all, this work proposes and investigates a semi-supervised learning method in forest inventory applications. It tackles four tree species classification problems while considering challenges like the low quality of satellite imagery and labeled data insufficiency. It also compares the proposed method to the conventional techniques of classification.

Firstly, let us compare the performance of SAE and CAE from Section \ref{exp_and_res_AE}. According to the results shown in Tables \ref{F1table} and \ref{sparse_ae_best_resuls1}, one can conclude that CAE suits our problem statement slightly better. However, since the differences between results are up to 0.3 in accuracy and f1 score, it is not excluded that a slightly different architecture or other ranges of parameters may make SAE superior. 

The main result comparing Tables \ref{acc2} and \ref{table3}, show that VGG has higher accuracy and f1 score than the considered 2D-CNN. Moreover, the VGG outperforms the SAE but is slightly worse than CAE. The 2D-CNN achieved the lowest scores of all considered models in our experiments and is therefore not applicable to our problem statement. In accordance with other authors, our results conclude that the SVC classifier fits this problem statement best. Followed by CatBoost, it achieved the best accuracy and f1 score in most cases.

From all experiments, we conclude that smaller image sizes lead to better classification results. Also, adding coordinates as additional features in the given data set and considering SAE does not improve accuracy and f1 score. Further, our experiments show that larger bottleneck sizes do not imply better results, which is counterintuitive.

For future investigations, CAEs and VGG model are the most promising approaches to our problem. Adding coordinates as additional features might increase the overall performance if the labeled data set grows and the coordinates get treated in classification methods different than the other features.

\newpage
%Bibliography
\bibliographystyle{unsrt}  
\bibliography{references}

\begin{thebibliography}{1}

\bibitem{heinzel_11}
J.~Heinzel and B.~Koch.
\newblock Exploring full-waveform lidar parameters for tree species
  classification.
\newblock {\em International Journal of Applied Earth Observation and
  Geoinformation}, 13(1):152 -- 160, 2011.

\bibitem{heinzel_12}
Johannes Heinzel and Barbara Koch.
\newblock Investigating multiple data sources for tree species classification
  in temperate forest and use for single tree delineation.
\newblock {\em International Journal of Applied Earth Observation and
  Geoinformation}, 18:101–110, 2012.

\bibitem{inproceedings}
Vasilii Mosin, Roberto Aguilar, Alexander Platonov, Albert Vasiliev, Alexander
  Kedrov, and Anton Ivanov.
\newblock Remote sensing and machine learning for tree detection and
  classification in forestry applications.
\newblock page~14, 2019.

\bibitem{Simonyan15}
Karen Simonyan and Andrew Zisserman.
\newblock Very deep convolutional networks for large-scale image recognition.
\newblock In {\em International Conference on Learning Representations}, 2015.

\bibitem{wang2017effectiveness}
Jason Wang and Luis Perez.
\newblock The effectiveness of data augmentation in image classification using
  deep learning.
\newblock {\em Convolutional Neural Networks Vis. Recognit}, page~11, 2017.

\end{thebibliography}

%\section*{Appendices}
\newpage
\section*{Appendix A \quad Convolutional Autoencoder}

%\renewcommand{\thesection}{Appendix\ \Alph{section}}
%\section{Convolutional Autoencoder}
\begin{table}[H]
\centering
\caption{Accuracy scores of the best classifiers for give bottleneck values and image size on the train/test set. Vertical columns stands for image size and horizontal one for bottleneck size.  \label{Accuracytable}
}
\begin{tabular}{|@{ }c@{ }|@{}c@{ }|@{}c@{ }|@{}c@{ }|@{}c@{ }|@{}c@{ }|}
\hline
&  \textbf{40} &   \textbf{60} &   \textbf{80} &   \textbf{100} &   \textbf{120} \\ \hline
  $\boldsymbol{10\times 10}$ & 
\begingroup\renewcommand{\arraystretch}{0.7} \begin{tabular}{@{}c@{}}   SVC:\\   0.96/0.753\end{tabular}\endgroup & \textbf{\begingroup\renewcommand{\arraystretch}{0.7} \begin{tabular}{@{}c@{}}  SVC:\\   0.97/0.768\end{tabular}\endgroup} & \begingroup\renewcommand{\arraystretch}{0.7} \begin{tabular}{@{}c@{}}  SVC: \\   0.97/0.743\end{tabular}\endgroup & \begingroup\renewcommand{\arraystretch}{0.7} \begin{tabular}{@{}c@{}}  CatB.:\\    0.99/0.763\end{tabular}\endgroup & \begingroup\renewcommand{\arraystretch}{0.7} \begin{tabular}{@{}c@{}}  SVC:\\    0.98/0.768\end{tabular}\endgroup \\ 
\hline
  $\boldsymbol{14\times 14}$ & 
\begingroup\renewcommand{\arraystretch}{0.7} \begin{tabular}{@{}c@{}}   SVC:\\   0.97/0.738\end{tabular}\endgroup & \begingroup\renewcommand{\arraystretch}{0.7} \begin{tabular}{@{}c@{}}   SVC:\\   0.98/0.768\end{tabular}\endgroup & \begingroup\renewcommand{\arraystretch}{0.7} \begin{tabular}{@{}c@{}}  SVC:\\   0.98/0.758\end{tabular}\endgroup & \begingroup\renewcommand{\arraystretch}{0.7} \begin{tabular}{@{}c@{}}  SVC:\\    0.97/0.748\end{tabular}\endgroup & \begingroup\renewcommand{\arraystretch}{0.7} \begin{tabular}{@{}c@{}}  SVC: \\   0.98/0.733\end{tabular}\endgroup \\ 
\hline
  $\boldsymbol{18\times 18}$ & 
\begingroup\renewcommand{\arraystretch}{0.7} \begin{tabular}{@{}c@{}}  SVC:\\    0.98/0.699\end{tabular}\endgroup & \begingroup\renewcommand{\arraystretch}{0.7} \begin{tabular}{@{}c@{}}  SVC: \\   0.98/0.753\end{tabular}\endgroup & \begingroup\renewcommand{\arraystretch}{0.7} \begin{tabular}{@{}c@{}}  SVC:\\    0.98/0.748\end{tabular}\endgroup & \begingroup\renewcommand{\arraystretch}{0.7} \begin{tabular}{@{}c@{}}  SVC:\\    0.98/0.714\end{tabular}\endgroup & \begingroup\renewcommand{\arraystretch}{0.7} \begin{tabular}{@{}c@{}}  SVC:\\    0.98/0.733\end{tabular}\endgroup \\ 
\hline
  $\boldsymbol{22\times 22}$ & 
\begingroup\renewcommand{\arraystretch}{0.7} \begin{tabular}{@{}c@{}}   SVC: \\    0.97/0.719\end{tabular}\endgroup & \begingroup\renewcommand{\arraystretch}{0.7} \begin{tabular}{@{}c@{}}  SVC:\\    0.98/0.729\end{tabular}\endgroup & \begingroup\renewcommand{\arraystretch}{0.7} \begin{tabular}{@{}c@{}}  SVC:\\   0.98/0.729\end{tabular}\endgroup & \begingroup\renewcommand{\arraystretch}{0.7} \begin{tabular}{@{}c@{}}  SVC:\\   0.98/0.743\end{tabular}\endgroup & \begingroup\renewcommand{\arraystretch}{0.7} \begin{tabular}{@{}c@{}}  SVC:\\    0.99/0.748\end{tabular}\endgroup \\ 
\hline
  $\boldsymbol{26\times 26}$ & 
\begingroup\renewcommand{\arraystretch}{0.7} \begin{tabular}{@{}c@{}}  SVC:\\    0.97/0.650\end{tabular}\endgroup & \begingroup\renewcommand{\arraystretch}{0.7} \begin{tabular}{@{}c@{}}  SVC:\\    0.98/0.689\end{tabular}\endgroup & \begingroup\renewcommand{\arraystretch}{0.7} \begin{tabular}{@{}c@{}}  SVC:\\   0.98/0.719\end{tabular}\endgroup & \begingroup\renewcommand{\arraystretch}{0.7} \begin{tabular}{@{}c@{}}  SVC:\\    0.99/0.743\end{tabular}\endgroup & \begingroup\renewcommand{\arraystretch}{0.7} \begin{tabular}{@{}c@{}}  SVC:\\    0.99/0.709\end{tabular}\endgroup \\ 
\hline
  $\boldsymbol{30\times 30}$ & 
\begingroup\renewcommand{\arraystretch}{0.7} \begin{tabular}{@{}c@{}}  SVC:\\    0.97/0.674\end{tabular}\endgroup & \begingroup\renewcommand{\arraystretch}{0.7} \begin{tabular}{@{}c@{}}  SVC:\\    0.98/0.669\end{tabular}\endgroup & \begingroup\renewcommand{\arraystretch}{0.7} \begin{tabular}{@{}c@{}}  SVC:\\    0.98/0.714\end{tabular}\endgroup & \begingroup\renewcommand{\arraystretch}{0.7} \begin{tabular}{@{}c@{}}  SVC:\\   0.99/0.724\end{tabular}\endgroup & \begingroup\renewcommand{\arraystretch}{0.7} \begin{tabular}{@{}c@{}}  SVC:\\    0.99/0.733\end{tabular}\endgroup \\ \hline
\end{tabular}
\end{table}
\begin{table}[H]
\centering
\caption{F1 scores of the best classifiers for give bottleneck values and image size on the train/test set. Vertical columns stands for image size and horizontal one for bottleneck size. \label{F7table}
}
\begin{tabular}{|@{ }c@{ }|@{}c@{ }|@{}c@{ }|@{}c@{ }|@{}c@{ }|@{}c@{ }|}
\hline
&  \textbf{40} &   \textbf{60} &   \textbf{80}&   \textbf{100}&   \textbf{120} 
\\ \hline
  $\boldsymbol{10\times 10}$ &
\begingroup\renewcommand{\arraystretch}{0.7} \begin{tabular}{@{}c@{}}   SVC:\\   0.96/0.636   \end{tabular}\endgroup &
\textbf{\begingroup\renewcommand{\arraystretch}{0.7} \begin{tabular}{@{}c@{}}  SVC:\\   0.97/0.645     \end{tabular}\endgroup} &
\begingroup\renewcommand{\arraystretch}{0.7} \begin{tabular}{@{}c@{}}  SVC: \\   0.97/0.631   \end{tabular}\endgroup &
\begingroup\renewcommand{\arraystretch}{0.7} \begin{tabular}{@{}c@{}}  SVC:\\   0.97/0.639 \end{tabular}\endgroup & \begingroup\renewcommand{\arraystretch}{0.7} \begin{tabular}{@{}c@{}}  SVC:\\   0.98/0.645 \end{tabular}\endgroup \\ 
\hline
  $\boldsymbol{14\times 14}$ & 
\begingroup\renewcommand{\arraystretch}{0.7} \begin{tabular}{@{}c@{}}   SVC:\\   0.97/0.571   \end{tabular}\endgroup & \begingroup\renewcommand{\arraystretch}{0.7} \begin{tabular}{@{}c@{}}   SVC:\\   0.98/0.6   \end{tabular}\endgroup & \begingroup\renewcommand{\arraystretch}{0.7} \begin{tabular}{@{}c@{}}  SVC:\\   0.98/0.59 \end{tabular}\endgroup & \begingroup\renewcommand{\arraystretch}{0.7} \begin{tabular}{@{}c@{}}  SVC:\\   0.97/0.588 \end{tabular}\endgroup & \begingroup\renewcommand{\arraystretch}{0.7} \begin{tabular}{@{}c@{}}  SVC: \\   0.91/0.581 \end{tabular}\endgroup \\ 
\hline
  $\boldsymbol{18\times 18}$ & 
\begingroup\renewcommand{\arraystretch}{0.7} \begin{tabular}{@{}c@{}}  SVC:\\   0.98/0.54 \end{tabular}\endgroup & \begingroup\renewcommand{\arraystretch}{0.7} \begin{tabular}{@{}c@{}}  SVC: \\   0.98/0.586\end{tabular}\endgroup & \begingroup\renewcommand{\arraystretch}{0.7} \begin{tabular}{@{}c@{}}  SVC:\\   0.98/0.575 \end{tabular}\endgroup & \begingroup\renewcommand{\arraystretch}{0.7} \begin{tabular}{@{}c@{}}  SVC:\\   0.98/0.556 \end{tabular}\endgroup & \begingroup\renewcommand{\arraystretch}{0.7} \begin{tabular}{@{}c@{}}  SVC:\\   0.99/0.574 \end{tabular}\endgroup \\ 
\hline
  $\boldsymbol{22\times 22}$ & 
\begingroup\renewcommand{\arraystretch}{0.7} \begin{tabular}{@{}c@{}}   SVC: \\   0.97/0.551 \end{tabular}\endgroup & \begingroup\renewcommand{\arraystretch}{0.7} \begin{tabular}{@{}c@{}}  SVC:\\   0.98/0.55 \end{tabular}\endgroup & \begingroup\renewcommand{\arraystretch}{0.7} \begin{tabular}{@{}c@{}}  SVC:\\   0.98/0.566 \end{tabular}\endgroup & \begingroup\renewcommand{\arraystretch}{0.7} \begin{tabular}{@{}c@{}}  SVC:\\   0.98/0.574 \end{tabular} \endgroup& \begingroup\renewcommand{\arraystretch}{0.7} \begin{tabular}{@{}c@{}}  SVC:\\   0.99/0.576 \end{tabular}\endgroup \\ 
\hline
  $\boldsymbol{26\times 26}$ & 
\begingroup\renewcommand{\arraystretch}{0.7} \begin{tabular}{@{}c@{}}  SVC:\\   0.97/0.509  \end{tabular}\endgroup & \begingroup\renewcommand{\arraystretch}{0.7} \begin{tabular}{@{}c@{}}  SVC:\\   0.98/0.536 \end{tabular}\endgroup & \begingroup\renewcommand{\arraystretch}{0.7} \begin{tabular}{@{}c@{}}  SVC:\\   0.98/0.555  \end{tabular}\endgroup & \begingroup\renewcommand{\arraystretch}{0.7} \begin{tabular}{@{}c@{}}  SVC:\\   0.99/0.577 \end{tabular}\endgroup & \begingroup\renewcommand{\arraystretch}{0.7} \begin{tabular}{@{}c@{}}  SVC:\\   0.99/0.551 \end{tabular}\endgroup \\ 
\hline
  $\boldsymbol{30\times 30}$ & 
\begingroup\renewcommand{\arraystretch}{0.7} \begin{tabular}{@{}c@{}}  SVC:\\   0.97/0.529 \end{tabular}\endgroup & \begingroup\renewcommand{\arraystretch}{0.7} \begin{tabular}{@{}c@{}}  SVC:\\   0.98/0.516 \end{tabular}\endgroup & \begingroup\renewcommand{\arraystretch}{0.7} \begin{tabular}{@{}c@{}}  SVC:\\    0.98/0.552  \end{tabular}\endgroup & \begingroup\renewcommand{\arraystretch}{0.7} \begin{tabular}{@{}c@{}}  SVC:\\  0.99/0.551 \end{tabular}\endgroup & \begingroup\renewcommand{\arraystretch}{0.7} \begin{tabular}{@{}c@{}}  SVC:\\   0.99/0.561  \end{tabular}\endgroup\\ 
\hline
\end{tabular}
\end{table}

\section*{Appendix B\quad Sparse Autoencoder}
\label{app_sparse_autoencoder}

%\begin{figure}[!htbp]
\begin{figure}[H]
% \vskip 0.2in
\begin{center}
\centerline{\includegraphics[width=0.7\columnwidth]{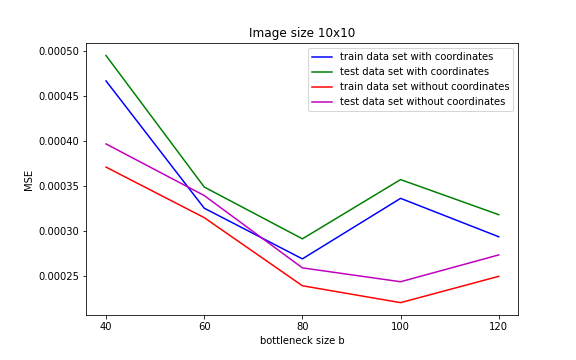}}
\caption{The mean squared error on train and test set after training the neural network with and without considering coordinates.}
\label{AE_MSE_s5}
\end{center}
\vskip -0.2in
\end{figure}

\begin{table}[H]
\centering
\caption{Highest achieved accuracy for combinations of bottleneck and image sizes using coordinates as additional features on the train/test set.
}
\begin{tabular}{|@{ }c@{ }|@{}c@{ }|@{}c@{ }|@{}c@{ }|@{}c@{ }|@{}c@{ }|}
\hline
&  \textbf{40} &  \textbf{60} &  \textbf{80} &  \textbf{100}&  \textbf{120} \\ 
\hline
  $\boldsymbol{10\times 10}$ & 
\begingroup\renewcommand{\arraystretch}{0.7} \begin{tabular}{@{}c@{}}   CatB.:\\   0.99/0.719\end{tabular}\endgroup & 
\textbf{\begingroup\renewcommand{\arraystretch}{0.7} \begin{tabular}{@{}c@{}}   CatB.: \\   0.99/0.739 \end{tabular}\endgroup} & 
\begingroup\renewcommand{\arraystretch}{0.7} \begin{tabular}{@{}c@{}}  CatB.: \\   0.99/0.719\end{tabular}\endgroup & 
\begingroup\renewcommand{\arraystretch}{0.7} \begin{tabular}{@{}c@{}}  CatB.: \\   0.99/0.734\end{tabular}\endgroup & 
\begingroup\renewcommand{\arraystretch}{0.7} \begin{tabular}{@{}c@{}}  CatB.: \\   0.99/0.714\end{tabular}\endgroup \\ 
\hline
  $\boldsymbol{14\times 14}$ & 
\begingroup\renewcommand{\arraystretch}{0.7} \begin{tabular}{@{}c@{}}  SVC: \\   0.93/0.734\end{tabular}\endgroup & 
\begingroup\renewcommand{\arraystretch}{0.7} \begin{tabular}{@{}c@{}}  CatB.:\\   0.99/0.685\end{tabular}\endgroup & 
\begingroup\renewcommand{\arraystretch}{0.7} \begin{tabular}{@{}c@{}}  CatB.: \\   0.99/0.714\end{tabular}\endgroup & 
\begingroup\renewcommand{\arraystretch}{0.7} \begin{tabular}{@{}c@{}}  CatB.: \\   0.99/0.729\end{tabular}\endgroup & 
\begingroup\renewcommand{\arraystretch}{0.7} \begin{tabular}{@{}c@{}}  CatB.: \\   0.99/0.714\end{tabular}\endgroup \\ 
\hline
  $\boldsymbol{18\times 18}$ & 
\begingroup\renewcommand{\arraystretch}{0.7} \begin{tabular}{@{}c@{}}  SVC: \\   0.95/0.685\end{tabular}\endgroup & 
\begingroup\renewcommand{\arraystretch}{0.7} \begin{tabular}{@{}c@{}}  RFC: \\   0.99/0.675\end{tabular}\endgroup & 
\begingroup\renewcommand{\arraystretch}{0.7} \begin{tabular}{@{}c@{}}  SVC: \\   0.99/0.7\end{tabular}\endgroup & 
\begingroup\renewcommand{\arraystretch}{0.7} \begin{tabular}{@{}c@{}}  SVC: \\   0.99/0.685 \end{tabular}\endgroup & 
\begingroup\renewcommand{\arraystretch}{0.7} \begin{tabular}{@{}c@{}}  CatB.: \\   0.99/0.69\end{tabular}\endgroup \\ 
\hline
  $\boldsymbol{22\times 22}$    & 
\begingroup\renewcommand{\arraystretch}{0.7} \begin{tabular}{@{}c@{}}  SVC: \\   0.96/0.69\end{tabular}\endgroup & 
\begingroup\renewcommand{\arraystretch}{0.7} \begin{tabular}{@{}c@{}}  CatB.: \\   0.99/0.685\end{tabular}\endgroup & 
\begingroup\renewcommand{\arraystretch}{0.7} \begin{tabular}{@{}c@{}}  CatB.: \\   0.99/0.655\end{tabular}\endgroup & 
\begingroup\renewcommand{\arraystretch}{0.7} \begin{tabular}{@{}c@{}}  CatB.: \\   0.99/0.65\end{tabular}\endgroup & 
\begingroup\renewcommand{\arraystretch}{0.7} \begin{tabular}{@{}c@{}}  CatB.: \\   0.99/0.67\end{tabular}\endgroup \\ 
\hline
  $\boldsymbol{26\times 26}$ & 
\begingroup\renewcommand{\arraystretch}{0.7} \begin{tabular}{@{}c@{}}  CatB.: \\   0.99/0.67\end{tabular}\endgroup & 
\begingroup\renewcommand{\arraystretch}{0.7} \begin{tabular}{@{}c@{}}  CatB.: \\   0.99/0.65\end{tabular}\endgroup & 
\begingroup\renewcommand{\arraystretch}{0.7} \begin{tabular}{@{}c@{}}  CatB.: \\   0.99/0.616\end{tabular}\endgroup & 
\begingroup\renewcommand{\arraystretch}{0.7} \begin{tabular}{@{}c@{}}  CatB.: \\   0.99/0.66\end{tabular}\endgroup & 
\begingroup\renewcommand{\arraystretch}{0.7} \begin{tabular}{@{}c@{}}  SVC: \\   0.99/0.67\end{tabular}\endgroup \\ 
\hline
  $\boldsymbol{30\times 30}$ & 
\begingroup\renewcommand{\arraystretch}{0.7} \begin{tabular}{@{}c@{}}  RFC: \\   0.99/0.631\end{tabular}\endgroup & 
\begingroup\renewcommand{\arraystretch}{0.7} \begin{tabular}{@{}c@{}}  CatB.: \\   0.99/0.635\end{tabular}\endgroup & 
\begingroup\renewcommand{\arraystretch}{0.7} \begin{tabular}{@{}c@{}}  CatB.: \\   0.99/0.67\end{tabular}\endgroup & 
\begingroup\renewcommand{\arraystretch}{0.7} \begin{tabular}{@{}c@{}}  CatB.: \\   0.99/0.635\end{tabular}\endgroup & 
\begingroup\renewcommand{\arraystretch}{0.7} \begin{tabular}{@{}c@{}}  SVC: \\   0.99/0.635\end{tabular}\endgroup \\ 
\hline
\end{tabular}
\end{table}

\begin{table}[H]
\centering
\caption{Highest achieved accuracy for combinations of bottleneck and image sizes without using coordinates as additional features on the train/test set.}
\begin{tabular}{|@{ }c@{ }|@{ }c@{ }|@{ }c@{ }|@{ }c@{ }|@{ }c@{ }|@{ }c@{ }|}
\hline
&   \textbf{40} &   \textbf{60} &   \textbf{80} &   \textbf{100}&   \textbf{120} \\ 
\hline

  $\boldsymbol{10\times 10}$ & 
\begingroup\renewcommand{\arraystretch}{0.7}\begin{tabular}{@{}c@{}}  CatB.:\\   0.99/0.739\end{tabular}\endgroup & 
\begingroup\renewcommand{\arraystretch}{0.7}\begin{tabular}{@{}c@{}}  CatB.:\\   0.99/0.724\end{tabular}\endgroup & 
\begingroup\renewcommand{\arraystretch}{0.7}\begin{tabular}{@{}c@{}}  CatB.:\\   0.99/0.724\end{tabular}\endgroup & 
\begingroup\renewcommand{\arraystretch}{0.7}\begin{tabular}{@{}c@{}}  CatB.:\\   0.99/0.714\end{tabular}\endgroup & 
\begingroup\renewcommand{\arraystretch}{0.7}\begin{tabular}{@{}c@{}}  SVC:\\   0.97/0.719\end{tabular}\endgroup \\ 
\hline

  $\boldsymbol{14\times 14}$ & 
\begingroup\renewcommand{\arraystretch}{0.7}\begin{tabular}{@{}c@{}}  CatB.:\\   0.99/0.704\end{tabular}\endgroup & 
\begingroup\renewcommand{\arraystretch}{0.7}\begin{tabular}{@{}c@{}}  CatB.:\\   0.99/0.719\end{tabular}\endgroup & 
\begingroup\renewcommand{\arraystretch}{0.7}\begin{tabular}{@{}c@{}}  CatB.:\\   0.99/0.734\end{tabular}\endgroup & 
\begingroup\renewcommand{\arraystretch}{0.7}\begin{tabular}{@{}c@{}}  CatB.:\\   0.99/0.719\end{tabular}\endgroup & 
\begingroup\renewcommand{\arraystretch}{0.7}\begin{tabular}{@{}c@{}}  SVC:\\   0.97/0.68\end{tabular}\endgroup \\ 
\hline
  $\boldsymbol{18\times 18}$ & 
\begingroup\renewcommand{\arraystretch}{0.7}\begin{tabular}{@{}c@{}}  CatB.: \\  0.99/0.695\end{tabular}\endgroup & 
\begingroup\renewcommand{\arraystretch}{0.7}\begin{tabular}{@{}c@{}}  CatB.: \\  0.99/0.665\end{tabular}\endgroup & 
\begingroup\renewcommand{\arraystretch}{0.7}\begin{tabular}{@{}c@{}}  RFC: \\  0.99/0.7\end{tabular}\endgroup & 
\begingroup\renewcommand{\arraystretch}{0.7}\begin{tabular}{@{}c@{}}  XGB: \\  0.99/0.709 \end{tabular}\endgroup & 
\begingroup\renewcommand{\arraystretch}{0.7}\begin{tabular}{@{}c@{}}  XGB: \\  0.99/0.709\end{tabular}\endgroup \\ \hline
  $\boldsymbol{22\times 22}$ & 
\begingroup\renewcommand{\arraystretch}{0.7}\begin{tabular}{@{}c@{}}  CatB.: \\   0.99/0.64\end{tabular}\endgroup & 
\begingroup\renewcommand{\arraystretch}{0.7}\begin{tabular}{@{}c@{}}  CatB.: \\   0.99/0.655\end{tabular}\endgroup & 
\begingroup\renewcommand{\arraystretch}{0.7}\begin{tabular}{@{}c@{}}  CatB.: \\   0.99/0.68\end{tabular}\endgroup & 
\begingroup\renewcommand{\arraystretch}{0.7}\begin{tabular}{@{}c@{}}  CatB.: \\   0.99/0.675\end{tabular}\endgroup & 
\begingroup\renewcommand{\arraystretch}{0.7}\begin{tabular}{@{}c@{}}  CatB.: \\   0.99/0.695\end{tabular}\endgroup \\ 
\hline
  $\boldsymbol{26\times 26}$ & 
\begingroup\renewcommand{\arraystretch}{0.7}\begin{tabular}{@{}c@{}}  RFC:\\    0.99/0.616\end{tabular}\endgroup & 
\begingroup\renewcommand{\arraystretch}{0.7}\begin{tabular}{@{}c@{}}  CatB.: \\   0.99/0.66\end{tabular}\endgroup & 
\begingroup\renewcommand{\arraystretch}{0.7}\begin{tabular}{@{}c@{}}  XGB: \\   0.99/0.66\end{tabular}\endgroup & 
\begingroup\renewcommand{\arraystretch}{0.7}\begin{tabular}{@{}c@{}}  CatB.: \\   0.99/0.665\end{tabular}\endgroup & 
\begingroup\renewcommand{\arraystretch}{0.7}\begin{tabular}{@{}c@{}}  CatB.: \\   0.99/0.66\end{tabular}\endgroup \\ 
\hline
  $\boldsymbol{30\times 30}$ & 
\begingroup\renewcommand{\arraystretch}{0.7}\begin{tabular}{@{}c@{}}  RFC: \\   0.99/0.596\end{tabular}\endgroup &
\begingroup\renewcommand{\arraystretch}{0.7}\begin{tabular}{@{}c@{}}  RFC: \\   0.99/0.645\end{tabular}\endgroup & 
\begingroup\renewcommand{\arraystretch}{0.7}\begin{tabular}{@{}c@{}}  CatB.: \\   0.99/0.665\end{tabular}\endgroup & 
\begingroup\renewcommand{\arraystretch}{0.7}\begin{tabular}{@{}c@{}}  CatB.: \\   0.99/0.645\end{tabular}\endgroup & 
\begingroup\renewcommand{\arraystretch}{0.7}\begin{tabular}{@{}c@{}}  XGB: \\   0.99/0.616\end{tabular}\endgroup \\ 
\hline
\end{tabular}
\end{table}

\begin{table}[H]
\centering
\caption{Highest achieved F1-score for combinations of bottleneck and image sizes using coordinates as additional features on the train/test set. \label{F8table}
}
\begin{tabular}{|@{ }c@{ }|@{ }c@{ }|@{ }c@{ }|@{ }c@{ }|@{ }c@{ }|@{ }c@{ }|}
\hline
&   \textbf{40} &   \textbf{60} &   \textbf{80} &   \textbf{100} &   \textbf{120} \\ 
\hline
  $\boldsymbol{10\times 10}$ & 
\begingroup\renewcommand{\arraystretch}{0.7}\begin{tabular}{@{}c@{}}  CatB.:\\   0.99/0.57\end{tabular}\endgroup  & 
\begingroup\renewcommand{\arraystretch}{0.7}\begin{tabular}{@{}c@{}}  CatB.: \\    0.99/0.565\end{tabular}\endgroup  & 
\begingroup\renewcommand{\arraystretch}{0.7}\begin{tabular}{@{}c@{}}  CatB.: \\   0.99/0.558\end{tabular}\endgroup  & 
\textbf{\begingroup\renewcommand{\arraystretch}{0.7}\begin{tabular}{@{}c@{}}  CatB.: \\   0.99/0.62\end{tabular}\endgroup } & 
\begingroup\renewcommand{\arraystretch}{0.7}\begin{tabular}{@{}c@{}}    CatB.: \\   0.99/0.548\end{tabular}\endgroup  \\ 
\hline
  $\boldsymbol{14\times 14}$ & 
\begingroup\renewcommand{\arraystretch}{0.7}\begin{tabular}{@{}c@{}}  SVC: \\   0.93/0.609\end{tabular}\endgroup  & 
\begingroup\renewcommand{\arraystretch}{0.7}\begin{tabular}{@{}c@{}}  CatB.: \\   0.99/0.529\end{tabular}\endgroup  & 
\begingroup\renewcommand{\arraystretch}{0.7}\begin{tabular}{@{}c@{}}  CatB.: \\   0.99/0.546\end{tabular}\endgroup  & 
\begingroup\renewcommand{\arraystretch}{0.7}\begin{tabular}{@{}c@{}}  CatB.: \\   0.99/0.61\end{tabular}\endgroup  & 
\begingroup\renewcommand{\arraystretch}{0.7}\begin{tabular}{@{}c@{}}  CatB.: \\   0.99/0.594\end{tabular}\endgroup  \\ 
\hline
  $\boldsymbol{18\times 18}$ & 
\begingroup\renewcommand{\arraystretch}{0.7}\begin{tabular}{@{}c@{}}  SVC: \\   0.95/0.536\end{tabular}\endgroup  & 
\begingroup\renewcommand{\arraystretch}{0.7}\begin{tabular}{@{}c@{}}  RFC: \\   0.99/0.507\end{tabular}\endgroup  & 
\begingroup\renewcommand{\arraystretch}{0.7}\begin{tabular}{@{}c@{}}  SVC: \\   0.99/0.538\end{tabular}\endgroup  & 
\begingroup\renewcommand{\arraystretch}{0.7}\begin{tabular}{@{}c@{}}  SVC: \\   0.99/0.523\end{tabular}\endgroup  & 
\begingroup\renewcommand{\arraystretch}{0.7}\begin{tabular}{@{}c@{}}  CatB.: \\   0.99/0.509\end{tabular}\endgroup \\ 
\hline
  $\boldsymbol{22\times 22}$ & 
\begingroup\renewcommand{\arraystretch}{0.7}\begin{tabular}{@{}c@{}}  SVC: \\   0.96/0.532\end{tabular}\endgroup  & 
\begingroup\renewcommand{\arraystretch}{0.7}\begin{tabular}{@{}c@{}}  CatB.: \\   0.99/0.555\end{tabular}\endgroup  & 
\begingroup\renewcommand{\arraystretch}{0.7}\begin{tabular}{@{}c@{}}  CatB.: \\   0.99/0.486\end{tabular}\endgroup  & 
\begingroup\renewcommand{\arraystretch}{0.7}\begin{tabular}{@{}c@{}}  CatB.: \\   0.99/0.486\end{tabular}\endgroup  & 
\begingroup\renewcommand{\arraystretch}{0.7}\begin{tabular}{@{}c@{}}  CatB.: \\   0.99/0.508\end{tabular}\endgroup  \\ 
\hline
  $\boldsymbol{26\times 26}$ & 
\begingroup\renewcommand{\arraystretch}{0.7}\begin{tabular}{@{}c@{}}  CatB.: \\   0.99/0.501\end{tabular}\endgroup  & 
\begingroup\renewcommand{\arraystretch}{0.7}\begin{tabular}{@{}c@{}}  CatB.: \\   0.99/0.489\end{tabular}\endgroup  & 
\begingroup\renewcommand{\arraystretch}{0.7}\begin{tabular}{@{}c@{}}  CatB.: \\   0.99/0.473\end{tabular}\endgroup  & 
\begingroup\renewcommand{\arraystretch}{0.7}\begin{tabular}{@{}c@{}}  CatB.: \\   0.99/0.492\end{tabular}\endgroup  & 
\begingroup\renewcommand{\arraystretch}{0.7}\begin{tabular}{@{}c@{}}  SVC:\\    0.99/0.564\end{tabular}\endgroup  \\ 
\hline
  $\boldsymbol{30\times 30}$ & 
\begingroup\renewcommand{\arraystretch}{0.7}\begin{tabular}{@{}c@{}}  RFC: \\   0.99/0.463\end{tabular}\endgroup  & 
\begingroup\renewcommand{\arraystretch}{0.7}\begin{tabular}{@{}c@{}}  CatB.: \\   0.99/0.491\end{tabular}\endgroup  & 
\begingroup\renewcommand{\arraystretch}{0.7}\begin{tabular}{@{}c@{}}  CatB.: \\   0.99/0.513\end{tabular}\endgroup  & 
\begingroup\renewcommand{\arraystretch}{0.7}\begin{tabular}{@{}c@{}}  CatB.: \\   0.99/0.493\end{tabular}\endgroup  & 
\begingroup\renewcommand{\arraystretch}{0.7}\begin{tabular}{@{}c@{}}  SVC: \\   0.99/0.498\end{tabular}\endgroup  \\
\hline
\end{tabular}
\end{table}

\begin{table}[H]
\centering
\caption{Highest achieved F1-score for combinations of bottleneck and image sizes without using coordinates as additional features on the train/test set.\label{F9table}
}
% \resizebox{\linewidth}{!}{%
\begin{tabular}{|@{ }c@{ }|@{ }c@{ }|@{ }c@{ }|@{ }c@{ }|@{ }c@{ }|@{ }c@{ }|}
\hline
&   \textbf{40} &   \textbf{60} &   \textbf{80} &   \textbf{100} &   \textbf{120} \\ \hline
  $\boldsymbol{10\times 10}$ & 
\textbf{\begingroup\renewcommand{\arraystretch}{0.7}\begin{tabular}{@{}c@{}}  CatB.: \\   0.99/0.62\end{tabular}\endgroup} & 
\begingroup\renewcommand{\arraystretch}{0.7}\begin{tabular}{@{}c@{}}  CatB.: \\   0.99/0.606\end{tabular}\endgroup & 
\begingroup\renewcommand{\arraystretch}{0.7}\begin{tabular}{@{}c@{}}  CatB.: \\   0.99/0.551\end{tabular}\endgroup & 
\begingroup\renewcommand{\arraystretch}{0.7}\begin{tabular}{@{}c@{}}  CatB.: \\   0.99/0.554\end{tabular}\endgroup & 
\begingroup\renewcommand{\arraystretch}{0.7}\begin{tabular}{@{}c@{}}  CatB.: \\   0.99/0.562\end{tabular}\endgroup \\ 
\hline
  $\boldsymbol{14\times 14}$ & 
\begingroup\renewcommand{\arraystretch}{0.7}\begin{tabular}{@{}c@{}}  CatB.: \\    0.99/0.539\end{tabular}\endgroup & 
\begingroup\renewcommand{\arraystretch}{0.7}\begin{tabular}{@{}c@{}}  CatB.: \\    0.99/0.541\end{tabular}\endgroup & 
\begingroup\renewcommand{\arraystretch}{0.7}\begin{tabular}{@{}c@{}}  CatB.: \\    0.99/0.608\end{tabular}\endgroup & 
\begingroup\renewcommand{\arraystretch}{0.7}\begin{tabular}{@{}c@{}}  CatB.: \\    0.99/0.551\end{tabular}\endgroup & 
\begingroup\renewcommand{\arraystretch}{0.7}\begin{tabular}{@{}c@{}}  SVC: \\    0.97/0.571\end{tabular}\endgroup \\ 
\hline
  $\boldsymbol{18\times 18}$ & 
\begingroup\renewcommand{\arraystretch}{0.7}\begin{tabular}{@{}c@{}}  CatB.: \\    0.99/0.523\end{tabular}\endgroup & 
\begingroup\renewcommand{\arraystretch}{0.7}\begin{tabular}{@{}c@{}}  CatB.: \\    0.99/0.504\end{tabular}\endgroup & 
\begingroup\renewcommand{\arraystretch}{0.7}\begin{tabular}{@{}c@{}}  RFC: \\    0.99/0.532\end{tabular}\endgroup & 
\begingroup\renewcommand{\arraystretch}{0.7}\begin{tabular}{@{}c@{}}  XGB: \\    0.99/0.537\end{tabular}\endgroup & 
\begingroup\renewcommand{\arraystretch}{0.7}\begin{tabular}{@{}c@{}}  XGB: \\    0.99/0.543\end{tabular}\endgroup \\ 
\hline
  $\boldsymbol{22\times 22}$ & 
\begingroup\renewcommand{\arraystretch}{0.7}\begin{tabular}{@{}c@{}}  CatB.: \\    0.99/0.484\end{tabular}\endgroup & 
\begingroup\renewcommand{\arraystretch}{0.7}\begin{tabular}{@{}c@{}}  CatB.: \\    0.99/0.505\end{tabular}\endgroup & 
\begingroup\renewcommand{\arraystretch}{0.7}\begin{tabular}{@{}c@{}}  CatB.: \\    0.99/0.513\end{tabular}\endgroup & 
\begingroup\renewcommand{\arraystretch}{0.7}\begin{tabular}{@{}c@{}}  CatB.: \\    0.99/0.518\end{tabular}\endgroup & 
\begingroup\renewcommand{\arraystretch}{0.7}\begin{tabular}{@{}c@{}}  CatB.: \\    0.99/0.528\end{tabular}\endgroup \\ 
\hline
  $\boldsymbol{26\times 26}$ & 
\begingroup\renewcommand{\arraystretch}{0.7}\begin{tabular}{@{}c@{}}  RFC: \\    0.99/0.447\end{tabular}\endgroup & 
\begingroup\renewcommand{\arraystretch}{0.7}\begin{tabular}{@{}c@{}}  CatB.: \\    0.99/0.502\end{tabular}\endgroup & 
\begingroup\renewcommand{\arraystretch}{0.7}\begin{tabular}{@{}c@{}}  XGB: \\    0.99/0.498\end{tabular}\endgroup & 
\begingroup\renewcommand{\arraystretch}{0.7}\begin{tabular}{@{}c@{}}  CatB.: \\    0.99/0.552\end{tabular}\endgroup & 
\begingroup\renewcommand{\arraystretch}{0.7}\begin{tabular}{@{}c@{}}  CatB.: \\    0.99/0.49\end{tabular}\endgroup \\ 
\hline
  $\boldsymbol{30\times 30}$ & \begingroup\renewcommand{\arraystretch}{0.7}\begin{tabular}{@{}c@{}}  RFC: \\    0.999/0.43\end{tabular}\endgroup & 
\begingroup\renewcommand{\arraystretch}{0.7}\begin{tabular}{@{}c@{}}  RFC: \\    0.99/0.468\end{tabular}\endgroup & 
\begingroup\renewcommand{\arraystretch}{0.7}\begin{tabular}{@{}c@{}}  CatB.: \\    0.99/0.506\end{tabular}\endgroup & 
\begingroup\renewcommand{\arraystretch}{0.7}\begin{tabular}{@{}c@{}}  CatB.: \\    0.99/0.489\end{tabular}\endgroup & 
\begingroup\renewcommand{\arraystretch}{0.7}\begin{tabular}{@{}c@{}}  XGB: \\    0.99/0.513\end{tabular}\endgroup \\ 
\hline
\end{tabular}
\end{table}

\section*{Appendix C \quad 2D--CNN}
\label{appendix-contrib}
\begin{table}[H]
\centering
\caption{Accuracy for every label, overall accuracy, and overall F1 score for all considered image sizes on the train/test set. Results in bold indicates the highest achieved value.\label{table2}}
\begin{tabular}{|@{ }c@{ }|@{ }c@{ }|@{ }c@{ }|@{ }c@{ }|@{ }c@{ }|@{ }c@{ }|@{ }c@{ }|}
\hline
 &   $\boldsymbol{10\times 10}$ &   $\boldsymbol{14\times 14}$ &   $\boldsymbol{18\times 18}$ &   $\boldsymbol{22\times 22}$ &   $\boldsymbol{26\times 26}$ &   $\boldsymbol{30\times 30}$  	 \\ 
 \hline 
  \textbf{Acc. Label 0} &   37.42  &   47.1 &   45.8 &   47.7 &   \textbf{52.3} &   49	 \\ 
\hline
  \textbf{Acc. Label 1}  &   \textbf{69.47} &   64.21 &   60 &   60.11 &   61.1 &   51.6 	 \\
\hline
  \textbf{Acc. Label 2}  &   69.23 &   61.54 &   \textbf{76.9} &   46.2 &   53.8 &   46.42  	 \\
\hline
  \textbf{Acc. Label 3}  &   92.68 &   92.68 &   \textbf{92.7} &   95.1 &   95.16 &   95.12 	 \\
\hline
  \textbf{Overall Acc.} &   56.25 &   60.3 &     57.89 &   58.55 &   \textbf{60.7} &   55.90 	 \\
\hline
  \textbf{Overall F1} &   0.55 &   \textbf{0.576} &   0.563 &   0.548 &   0.56 &   0.53 	 \\
\hline 
\end{tabular}
\end{table}

\begin{figure}[H]
\centering
\includegraphics[width=0.7\linewidth]{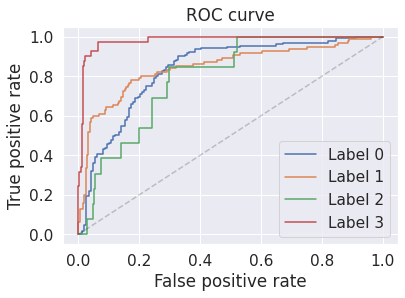}
\caption{Receiver Operative Characteristic (ROC) curve of the CNN model.  \label{fig44}}
\end{figure}

\begin{figure}[H]
\centering
\includegraphics[width=0.45\linewidth]{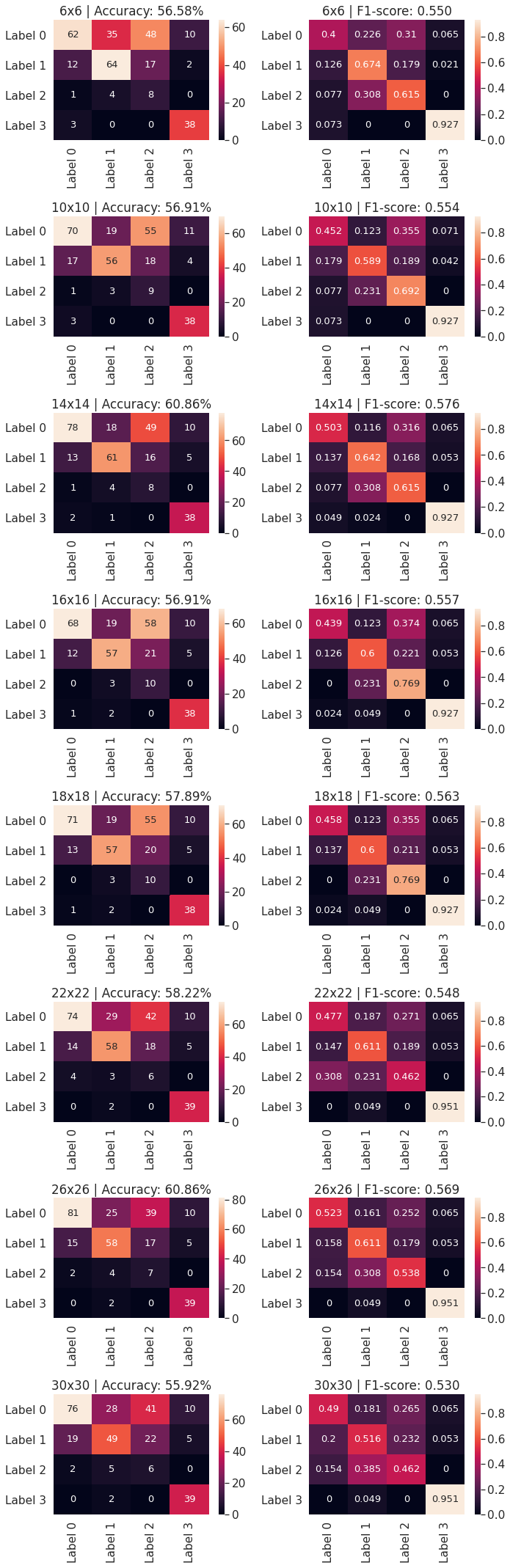}
\caption{ Confusion Matrices for 2D-CNN for each image size.  \label{fig45}}
\end{figure}

\section*{Appendix D \quad VGG and Upgraded 2D-CNN}
\label{appendix-contrib}

\subsection{VGG}
\label{Vgg_net}
The first block of the used VGG network has the following parameters: 
\begin{itemize}
\item Conv1(8, 32, kernel, padding=1), 
\item Conv2(32, 64, kernel=2, padding=1), 
\item MaxPool2d(kernel, stride=2), 
\item Conv3(64, 128, kernel=2, padding=1), 
\item Conv4(128, 128, kernel=2, padding=1), 
\item MaxPool2d(kernel, stride=2), 
\item Conv5(128, 256, kernel=2, padding=1), 
\item Conv6(256, 256, kernel=2, padding=1), 
\item second block has parameters Linear($x$, 4096),
\item Dropout(p=0.05), Linear(4096, 1024), 
\item Dropout(p=0.5), Linear(1024,128), Linear(128,4), 
\end{itemize}
where $x$ represents the image size. 

\subsection*{Upgraded 2D-CNN}
\label{custom_net_}
This CNN consists of two blocks of fully connected networks. The architecture is shown in Figure \ref{fig46}. We tried to avoid overfitting by using Dropout layers and a smaller network architecture. The test dataset was also balanced by the duplicating. The horizontal and vertical 'flip' procedures were performed, each with a probability of 0.5. The first block has the following parameters: 

\begin{itemize}
     \item Conv1(8, 32, kernel = 3, stride = 1, padding =1), 
     \item Conv2(32, 32, kernel = 3, stride = 1, padding =1), Pool(kernel = 2, stride = 2, padding = 0),
     \item DropOut(p=0.25), 
\end{itemize}
The second block has the following parameters: 
\begin{itemize}
\item Conv3(32, 64, kernel = 3, stride = 1, padding =1),
\item Conv4(64, 64, kernel = 3, stride = 1, padding =1),
\item Pool(kernel = 2, stride = 2, padding = 0),
\item DropOut(p=0.25),
\end{itemize}
and the last block has parameters 
\begin{itemize} 
\item Linear $(64x^2, 512)$, 
\item DropOut(p=0.5), Linear(512, 4), 
 \end{itemize}
 where $x$ represents the image size. 
 
\end{document}